\documentclass[sigconf]{acmart}
\usepackage{tcolorbox}
\usepackage{appendix}
\usepackage{subcaption}
\AtBeginDocument{%
  }

\copyrightyear{2025}
\acmYear{2025}
\setcopyright{cc}
\setcctype{by}
\acmConference[MM '25]{Proceedings of the 33rd ACM International Conference on Multimedia}{October 27--31, 2025}{Dublin, Ireland}
\acmBooktitle{Proceedings of the 33rd ACM International Conference on Multimedia (MM '25), October 27--31, 2025, Dublin, Ireland}
\acmDOI{10.1145/3746027.3758222}
\acmISBN{979-8-4007-2035-2/2025/10}

\usepackage{multirow}

\begin{document}

\title{SynthVLM: Towards High-Quality and Efficient Synthesis of Image-Caption Datasets for Vision-Language Models}

\author{Zheng Liu$^{\spadesuit}$}
\affiliation{%
  \institution{Peking University}
  \city{Beijing}
  \country{China}}
\email{2501213330@stu.pku.edu.cn}

\author{Hao Liang$^{\spadesuit}$}
\affiliation{%
  \institution{Peking University}
  \city{Beijing}
  \country{China}}
\email{hao.liang@stu.pku.edu.cn}

\author{Bozhou Li}
\affiliation{%
  \institution{Peking University}
  \city{Beijing}
  \country{China}}
\email{2301213084@pku.edu.cn}

\author{Wentao Xiong}
\affiliation{%
  \institution{Peking University}
  \city{Beijing}
  \country{China}}
\email{wtxiong@pku.edu.cn}

\author{Chong Chen}
\affiliation{%
  \institution{Huawei Technologies Ltd.}
  \city{Beijing}
  \country{China}}
\email{chenchong55@huawei.com}

\author{Conghui He}
\affiliation{%
  \institution{Shanghai AI Laboratory}
  \city{Shanghai}
  \country{China}}
\email{heconghui@pjlab.org.cn}

\author{Wentao Zhang$^\dagger$}
\affiliation{%
  \institution{Peking University}
  \department{Center for Machine Learning Research}
  \city{Beijing}
  \country{China}}
\email{wentao.zhang@pku.edu.cn}

\author{Bin Cui$^\dagger$}
\affiliation{%
  \institution{Peking University}
  \department{School of Computer Science \& Key Lab of High Confidence Software Technologies (MOE)}
  \city{Beijing}
  \country{China}}
\email{bin.cui@pku.edu.cn}


\renewcommand{\authors}{Zheng Liu, Hao Liang, Bozhou Li, Wentao Xiong, Chong Chen, Conghui He, Wentao Zhang, Bin Cui}
\renewcommand{\shortauthors}{liu et al.}

\begin{abstract}
Vision-Language Models (VLMs) have recently emerged, demonstrating remarkable vision-understanding capabilities. However, training these models requires large-scale datasets, which brings challenges related to efficiency, effectiveness, and quality of web data. In this paper, we introduce SynthVLM, a new data synthesis and curation method for generating image-caption pairs. Unlike traditional methods, where captions are generated from images, SynthVLM utilizes advanced diffusion models and high-quality captions to synthesize and select images from text captions, thereby creating precisely aligned image-text pairs. We further introduce SynthVLM-100K, a high-quality dataset consisting of 100K curated and synthesized image-caption pairs. In both model and human evaluations, SynthVLM-100K outperforms traditional real-world datasets. Leveraging this dataset, we develop a new family of multimodal large language models (MLLMs), SynthVLM-7B and SynthVLM-13B, which achieve state-of-the-art (SOTA) performance on various vision question-answering (VQA) tasks. Notably, our models outperform LLaVA across most metrics with only 18\% pretrain data. Furthermore, SynthVLM-7B and SynthVLM-13B attain SOTA performance on the MMLU benchmark, demonstrating that the high-quality SynthVLM-100K dataset preserves language abilities. Our dataset and the complete data generating and curating methods can be found in \url{https://github.com/starriver030515/SynthVLM}. 
\end{abstract}






\maketitle

\begingroup
\renewcommand\thefootnote{}\footnote{\noindent
$\spadesuit$ The first two authors have equal contributions. \\
$\dagger$ Corresponding Author
}
\addtocounter{footnote}{-1}
\endgroup

\begin{figure}
\centering 
\includegraphics[width=0.48\textwidth]{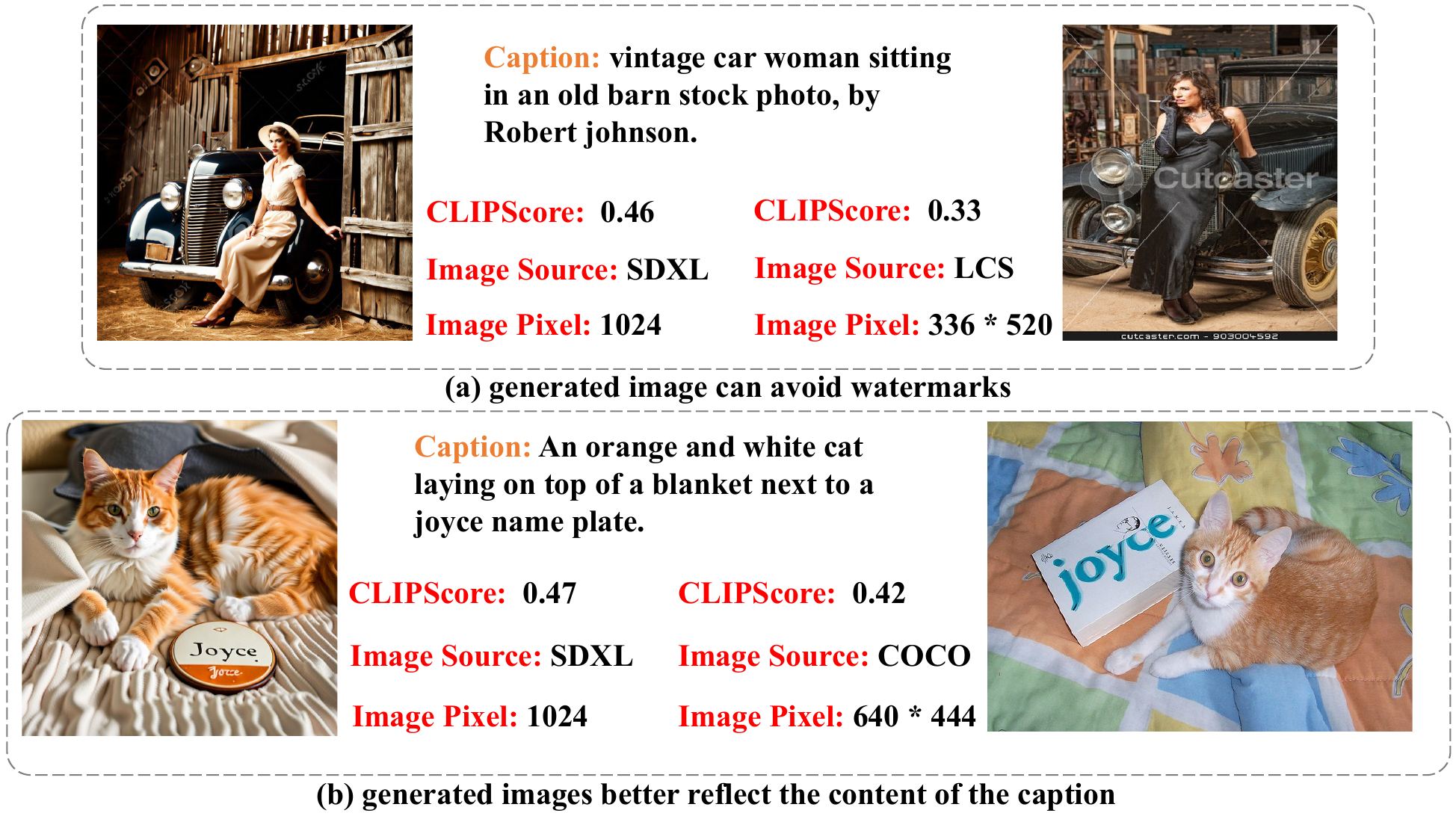} 
\caption{We compared SynthVLM-100K with LLaVA-558K. In (a), generated image can avoid content such as watermarks and advertisements. In (b), the generated images better reflect the content of the captions. Additionally, the resolution of the generated images is higher than real images.}
\label{fig: Face_1}
\end{figure}

\section{Introduction}
In recent years, with the rapid advancements in large language models (LLMs)~\cite{chatgpt, llama} and multimodal large language models (MLLMs)~\cite{zhao2023survey,wu2023multimodal}, data management has become a crucial aspect of these technologies~\cite{fernandez2023large, trummer2023bert, chen2023lingua, miao2024demystifying, nie2023flexmoe,cross-m}. At the same time, ~\cite{bai2024survey, Relational_Data, DB-GPT, Blockchains_and_Databases} also demonstrates that data processing, selection, and management can significantly influence the performance of MLLMs. 

\begin{figure*}[ht]
  \includegraphics[width=0.9\textwidth,height=0.41\textwidth]{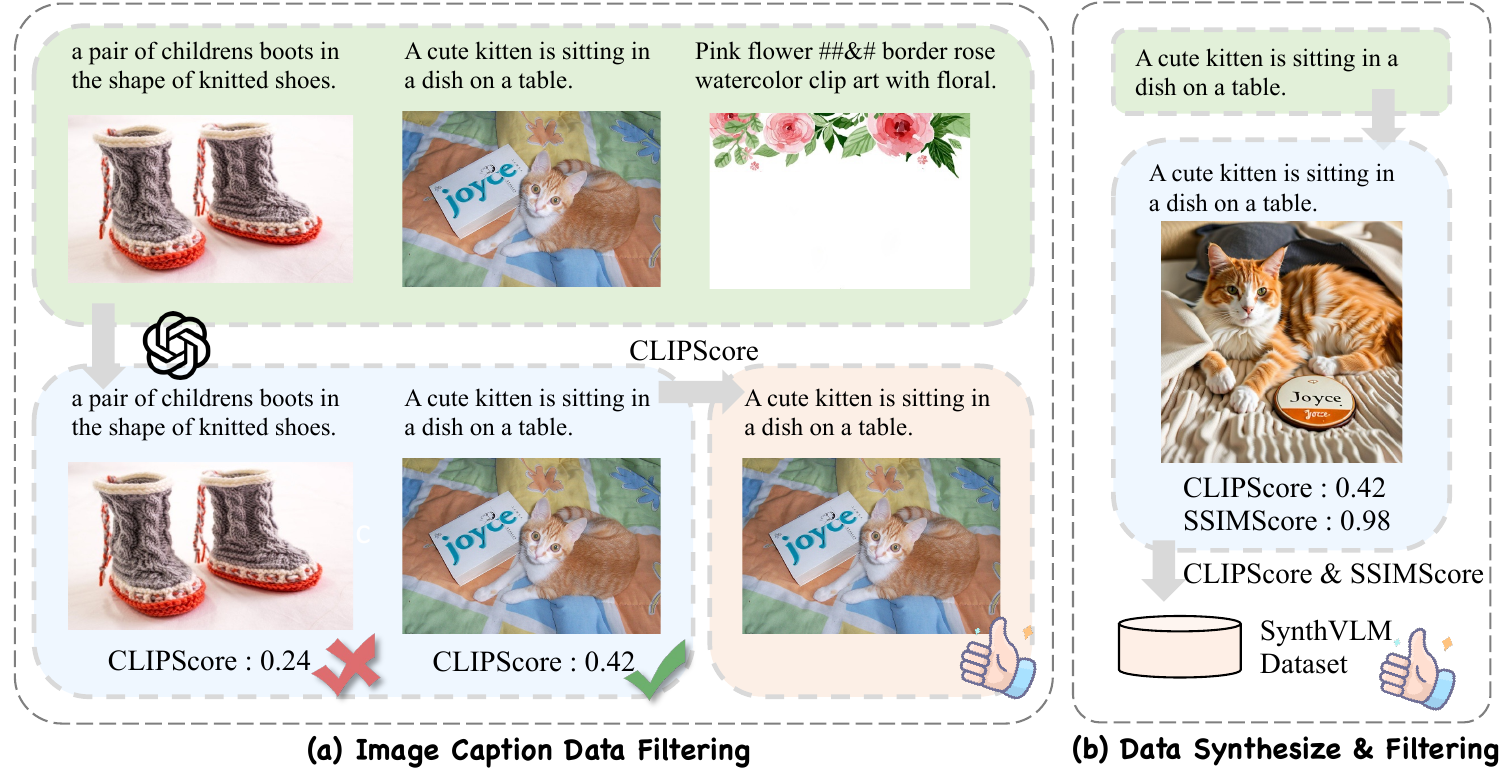}
  \caption{The pipeline of the SynthVLM data synthesis method is as follows: First, we filter high-quality image-caption pairs. Next, we synthesize high-quality data and subsequently filter them based on CLIPScore.}
  \label{fig: Main_Algorithm}
\end{figure*}

Vision Language Models (VLMs), a subset of MLLMs, excel in tasks like image classification, understanding, and captioning~\cite{chen2024internvl, blip, li2023blip, bai2023qwen}. While most VLMs focus on improving model architectures to integrate multimodal data~\cite{blip, li2023blip, bai2023qwen, llava, llava1.5, chen2024internvl}, their reliance on web-sourced data poses a bottleneck: high-quality, precisely aligned image-text pairs are scarce. Noisy or misaligned data directly limits performance, as shown by ~\cite{videochat2, internvideo2}, who link enhanced VLM effectiveness to carefully curated datasets. Crucially, precise alignment between modalities underpins data quality, which is a challenge demanding greater focus than architectural refinements alone.

To address the scarcity of high-quality web data, data synthesis strategies have increasingly been employed~\cite{DTIA, imagecaption, sharegpt4v}. For instance, ~\cite{imagecaption} utilized BLIP2 to generate numerous image captions, achieving SOTA results on DataComp. In the domain of VLMs, ~\cite{sharegpt4v} used GPT-4 Vision to produce highly descriptive image captions, resulting in significant improvements in LLaVA. The integration of these generative models has opened new avenues for enhancing data quality and alignment, further boosting VLM performance.

Despite these notable advancements in VLMs, the challenge of multimodal data persists, particularly when paired image and text are required. Although synthetic captions have been generated for images, the absence of generated images means that the issue of limited high-quality web images remains unresolved. This limitation gives rise to three key challenges that still need to be addressed:

\paragraph{\textbf{C1. Low Data Quality.}}Web-sourced images often contain artifacts such as blurriness and watermarks, compromising quality. Furthermore, caption generation approaches~\cite{imagecaption} leveraging BLIP2 tend to introduce logical inconsistencies and grammatical errors in text, which propagate through VLM training pipelines, ultimately weakening their linguistic reasoning capabilities.

\paragraph{\textbf{C2.Poor Effectiveness.}}
Existing datasets frequently lack the diversity and depth required to train VLMs effectively. Many web-scraped datasets consist of low-quality or irrelevant content that fails to capture the complexity of natural language and visual information. Furthermore, the limited scope of current datasets often results in models that struggle to generalize across various contexts.

\paragraph{\textbf{C3. Low Efficiency.}}
Methods that rely on manual captioning are both labor-intensive and resource-intensive. Automated solutions such as ShareGPT4V~\cite{sharegpt4v}, which leverage GPT-4 for labeling, are costly and difficult to scale. Additionally, many strategies require the creation of large datasets, resulting in significant data redundancy.

To address these challenges, we introduced a new data generation pipeline, SynthVLM. First, we implemented a quality selection process to filter high-quality caption data. Next, we employed diffusion models to generate images based on these captions. For quality assurance, we proposed a new method for evaluating and selecting image-caption pairs by combining CLIPScore~\cite{clipscore} and SSIM~\cite{ssim}, focusing on both image-text alignment and image quality. Our data generation approach achieved superior alignment between images and captions compared to existing methods. With 100K curated synthetic data, we attained SOTA results across multiple benchmarks, utilizing only 18\% of the official LLaVA-558K dataset size. Overall, our contributions are as follows:

\begin{itemize}
\item \textbf{Innovative Data Synthesis Framework.}
We introduce SynthVLM, a novel two-stage pipeline for generating synthetic image-caption pairs, and curate SynthVLM-100K—a large-scale, high-quality dataset produced through this framework. Compared to conventional approaches, SynthVLM achieves superior efficiency without compromising quality.

\item \textbf{Superior Synthetic Data Quality.}
SynthVLM-100K outperforms baseline datasets in both model and human evaluations. It achieves superior CLIPScore and SSIMScore, while also earning higher human evaluation ratings. Further validation via GPT-4 and Intern-VL2 assessments corroborates its enhanced quality for training vision-language models.

\item \textbf{State-of-the-Art Vision-Language Modeling.}
When pretrained exclusively on SynthVLM-100K, our models surpass baseline methods relying on LLaVA-558K, achieving top-tier performance on vision-language understanding benchmarks. SynthVLM-13B also delivers exceptional performance on the MMLU benchmark, highlighting its multimodal versatility.

\end{itemize}

\section{Related Work}
\subsection{Data Quality and Selection}
High-quality data can significantly enhance the performance of models~\cite{llama3repo}. Ensuring high data quality becomes more challenging because it requires more resources for data cleaning, selection and annotation~\cite{bai2024survey}. LLMs-based methods were commonly used in data selection~\cite{bai2024survey}. For instance, ~\cite{du2023mods} leverages DeBERTa~\cite{he2020deberta} for scoring, retaining high-quality data, and combining it with the k-center greedy algorithm to select diverse data. ~\cite{chen2023alpagasus} scores the accuracy of data using ChatGPT to pick out high-quality data. ~\cite{xu2023rethinking} use GPT-4 to rewrite data to increase their complexity and then streamline it by reducing its variety and improving its quality. 

\subsection{Data Generation}
Recent advancements in generating synthetic data and improving the performance of LLMs have shown promising results across various domains. A key component in generating high-quality synthetic datasets is precise alignment. ~\cite{reformattedalign} introduce REALIGN, a method that enhances the quality of instruction data by reformatting responses to better align with pre-established criteria and evidence, thereby improving LLMs' alignment with human values while minimizing human annotation and model hallucinations. ~\cite{selfalignment} build a high-quality instruction-following language model by automatically labeling human-written text with corresponding instructions and demonstrating highly effective self-alignment. 

\section{Method}

\subsection{Step1: Synthetic Dataset Construction}
\label{sec: synthetic_construction}
In this section, we introduce the image generation pipeline. First, we construct a large pool of captions. We then select the best captions from the pool for image-text generation. Utilizing these high-quality captions, we employ diffusion models to generate the images.

\paragraph{\textbf{Data Source.}}
To ensure the diversity of the captions, we combined human-generated and model-generated captions. As shown in  Table ~\ref{table: data_intro}. The human-generated captions were primarily sourced from LAION, CC, and SBU, while the model-generated captions were created using the method described in ~\cite{imagecaption}, which utilizes BLIP2 to regenerate captions for images in the DataComp dataset ~\cite{datacomp}. 

\paragraph{\textbf{Caption Curation.}}
To maintain dataset quality, we first removed low-quality captions, such as advertisements, overly repetitive descriptions, and captions with significant grammatical errors. The filtering process was performed with ChatGPT, combined with statistical indicators such as N-grams and Perplexity, ensuring that only high-quality, informative captions were used for training. 
\begin{table}
\caption{LCS abbreviates the LAION, CC, and SBU datasets. SynthVLM uses captions to generate images, while others use images to generate captions or manual labeling.}
\resizebox{0.48\textwidth}{!}{
    \centering
    \label{Table 1}
    \begin{tabular}{c | ccc}
    \toprule
    \textbf{Name} & \textbf{Image Source} & \textbf{Caption Source} & \textbf{Sample} \\
    \midrule
    COCO-Caption~\cite{coco} & COCO & Human & 118K \\
    BLIP-LCS~\cite{llava1.5} & LCS & BLIP & 558K \\
    ShareGPT4V~\cite{sharegpt4v} & LCS, COCO, etc & GPT4-Vision & 100K \\
    ShareGPT4V-PT~\cite{sharegpt4v} & LCS, COCO, etc & Share-Captioner & 1246K \\
    \midrule
    SynthVLM & Diffusion & LCS, COCO, BLIP2-DataComp, etc & 1000K \\
    \bottomrule
    \end{tabular}
}
\label{table: data_intro}
\end{table}
For the remaining captions, we calculated the CLIPScore\cite{clipscore} for these captions and their corresponding raw images. CLIPScore is a metric that measures the cosine similarity between images and their corresponding captions. The formula for calculating CLIPScore is as follows:
\begin{equation}\label{eq:CLIPScore}
    {CLIPScore}(I, C) = \frac{{CLIP}(I) \cdot {CLIP}(C)}{|| {CLIP}(I)|| \cdot || {CLIP}(C)||}
\end{equation}
where \( I \) represents the image, \( C \) represents the caption, and \( {CLIP}(I) \) and \( {CLIP}(C) \) denote the image and text feature vectors extracted by the CLIP model. The dot product of the vectors is denoted by \( \cdot \), and \( ||\cdot|| \) denotes the norm of the vectors.

We selected the top 40\% of image-caption pairs with the highest CLIPScores. These selected captions were included in the candidate caption set. Ultimately, we curated a dataset of 1M captions for data generation. By using only captions, our method significantly reduces storage overhead and processing time. The caption curation pipeline is summarized in Figure \ref{fig: Main_Algorithm}(a).

\paragraph{\textbf{Image Generation.}}
After filtering 1M high-quality captions, we employed Stable Diffusion XL (SDXL) ~\cite{sdxl}, a SOTA model capable of efficiently generating high-quality, high-resolution images. SynthVLM produces images at a resolution of 1024x1024, effectively addressing the low-resolution issues present in existing datasets. This improvement greatly enhances the quality and utility of the training data across various image generation and recognition tasks.

\begin{table*}[htbp]
\centering
\caption{Comparison of SynthVLM and LLaVA using the same model structure. We can see SynthVLM outperforms LLaVA on all the evaluation benchmarks.}
\resizebox{1.0\textwidth}{!}{
    \begin{tabular}{ccccccccccccc}
    \toprule
    \textbf{Models} & \textbf{LLM} & \textbf{SQA} & \textbf{SQA$^I$} & \textbf{MMVet} & \textbf{VizWiz} & \textbf{VQAv2} & \textbf{GQA} & \textbf{MMB} & \textbf{MME$^P$} & \textbf{MME$^C$} & \textbf{PoPE} & \textbf{MMLU} \\
    \midrule
    \textbf{LLaVA-7B}  &  Vicuna-1.5-7B    & 69.3 & 67.3  & 30.5  & \textbf{49.9} & 78.7 & 62.5 & 65.3 & 1484.8 & 315.6 & 86.0 & 36.3 \\
    \textbf{SynthVLM-7B}  & Vicuna-1.5-7B     & \textbf{70.4} & \textbf{68.9} & \textbf{32.2}  & 49.3 & \textbf{79.4} & \textbf{63.1} & \textbf{66.8} & \textbf{1518.5} & \textbf{345.7} & \textbf{87.0} & \textbf{41.2} \\
    \midrule
    \textbf{LLaVA-13B}  &  Vicuna-1.5-13B    & 74.2 & 71.0  & 35.0  & 53.6 & 80.0 & 63.0 & 67.7 & 1531.3 & 294.5 & 86.9 & 52.4 \\
    \textbf{SynthVLM-13B}  & Vicuna-1.5-13B     & \textbf{74.9} & \textbf{72.5} & \textbf{35.0}  & \textbf{55.9} & \textbf{80.0} & \textbf{63.5} & \textbf{68.3} & \textbf{1573.0} & \textbf{316.1}  & \textbf{88.4} & \textbf{54.6} \\
    \bottomrule
    \end{tabular}
}
\label{tab: Main_1}
\end{table*}

\begin{table*}[ht]
\centering
\caption{Result comparison of MMLU shows that with the synthetic 100k data, our SynthVLM outperforms LLaVA in pure language tasks. This demonstrates the effectiveness of the synthetic data in modality alignment.}
\begin{tabular}{@{}cccc|cccc@{}} 
\toprule 

\multirow{2}{*}{\textbf{Models}} & \multirow{2}{*}{\textbf{LLM}} & \multirow{2}{*}{\textbf{SQA}} & \multicolumn{5}{c}{\textbf{MMLU}} \\ 
\cmidrule(lr){4-8} 
& & & \textbf{Avg} & \textbf{STEM} & \textbf{Humanities} & \textbf{Social Sciences} & \textbf{Other} \\
\midrule 
\textbf{LLaVA-7B} & Vicuna-1.5-7B & 69.3 & 36.3 & 28.6 & 33.4 & 39.5 & 44.5 \\
\textbf{SynthVLM-7B} & Vicuna-1.5-7B & \textbf{70.4} & \textbf{41.2} & \textbf{31.7} & \textbf{37.4} & \textbf{47.0} & \textbf{50.2} \\
\midrule
\textbf{LLaVA-13B} & Vicuna-1.5-13B & 74.2 & 52.4 & 41.9 & 45.8 & 62.9 & 61.8 \\
\textbf{SynthVLM-13B} & Vicuna-1.5-13B & \textbf{74.9} & \textbf{54.6} & \textbf{45.0} & \textbf{49.3} & \textbf{64.0} & \textbf{62.2} \\
\bottomrule 
\end{tabular}
\label{tab: Main_2}
\end{table*}
\subsection{Step2: Synthetic Data Selection}\label{sec: synthetic_selection}
In this section, we introduce a novel algorithm for quality control of generated datasets. To better ensure the alignment between images and their corresponding text descriptions, we continue to use CLIPScore. For a given image \( I \) and its corresponding text \( C \), we first calculate the image-text CLIPScore, \({CLIPScore}(I, C) \), to assess their alignment. Additionally, the generated images have a resolution of \( 1024 \times 1024 \), which will be resized to \( 336 \times 336 \) for compatibility with CLIP~\cite{clip}. To account for the potential loss in quality due to this resizing process, we introduce the Structural Similarity Index Measure (SSIM), a metric for image quality control. This will help us better ensure the image quality is preserved after resizing. The SSIM formula is defined as follows:
\[
SSIM(x, y) = \frac{(2\mu_x \mu_y + C_1)(2\sigma_{xy} + C_2)}{(\mu_x^2 + \mu_y^2 + C_1)(\sigma_x^2 + \sigma_y^2 + C_2)}
\]

Where \( x \) and \( y \) are the two images being compared, \( \mu_x \) and \( \mu_y \) are the mean pixel intensities of the images, \( \sigma_x^2 \) and \( \sigma_y^2 \) are the variances of the images, \( \sigma_{xy} \) is the covariance of the images, \( C_1 \) and \( C_2 \) are constants to stabilize the division with weak denominator values.

For a given image \( I \) with a resolution of \( 1024 \times 1024 \), we first resize it to \( 336 \times 336 \), then interpolate to restore it back to \( 1024 \times 1024 \). The SSIM value is then computed between the resized image and the original image to quantify the loss introduced by the resizing process. Let \( I_{{resized}} \) denote the resized image, where:
\[
I_{{resized}} = {resize}({resize}(I, (336,336)), (1024,1024))
\]

The SSIMScore is computed as:
\[
SSIMScore = SSIM(I, I_{{resized}})
\]

\begin{table}
\caption{We compared the average CLIPScore, SSIMScore, and their weighted score across our synthetic dataset, ShareGPT4V, COCO-Caption, and BLIP-LCS. The results demonstrate that SynthVLM achieves the highest alignment and image quality.}
\resizebox{0.47\textwidth}{!}{
    \centering
    \begin{tabular}{c | c  c  c  c }
    \toprule
    \textbf{Datasets} & \textbf{\# Samples} & \textbf{CLIPScore} & \textbf{SSIMScore} & \textbf{Weighted\_Score}  \\
    \midrule
    COCO-Caption & 118K & 0.31 & 0.73 & 0.67\\
    BLIP-LCS & 558K & 0.32 & 0.75 & 0.70\\
    ShareGPT4V & 100K & 0.32 & 0.79 & 0.71\\
    \midrule
    Synth Dataset & 1000K & 0.34 & 0.78 & 0.73\\
    Curated-Synth Dataset & 100K & \textbf{0.36} & \textbf{0.86} & \textbf{0.79}\\
    \bottomrule
    \end{tabular}
}
\label{table: CLIPScore}
\end{table}

Finally, to combine the CLIPScore and SSIMScore, we apply a weighted sum, with SSIMScore weighted by a factor \( \lambda \). The overall score \( S \) is given by:
\[
Weighted\_Score = {CLIPScore}(I, C) + \lambda \cdot SSIMScore(I, I_{{resized}})
\]

In practice, we set \( \lambda = 0.5 \) to balance the contribution of CLIPScore and SSIMScore in the final score.

As shown in Figure \ref{fig: Main_Algorithm}(b), we initially computed CLIPScores and SSIMScore for the 1M synthetic image-caption pairs. We then selected the top 100K pairs that demonstrated the highest scores, indicating the most accurate and meaningful matches between images and captions. By curating this subset, we constructed a high-quality, highly aligned synthetic dataset. 

\begin{figure}[t]
\centering 
\includegraphics[width=0.47\textwidth]{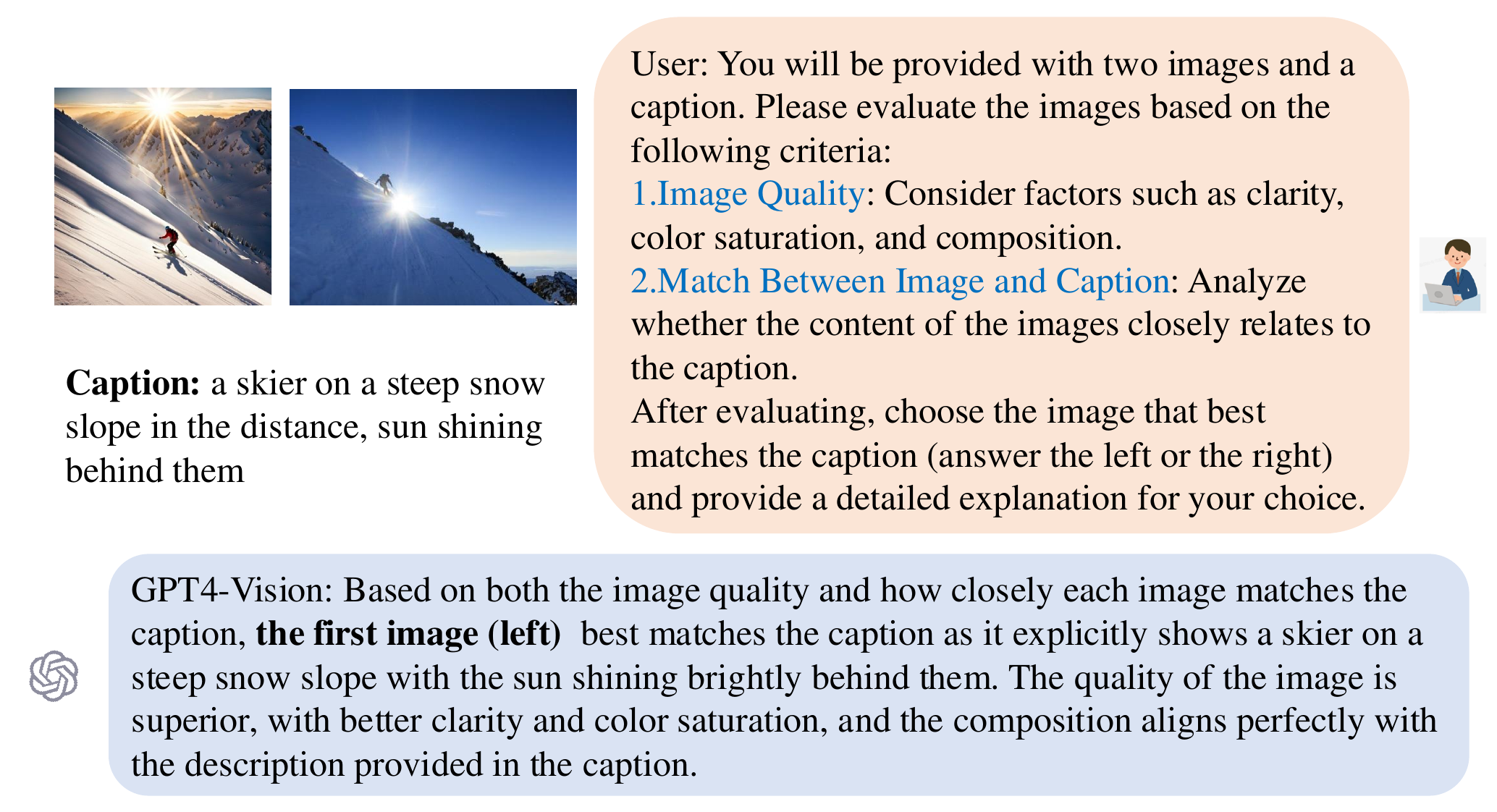} 
\caption{Our process and prompt design for match assessment using GPT4V. We consider various aspects, including the quality of the image and the match between the image and the caption. Based on this process, we compare SynthVLM with existing datasets from the model's perspective.}
\label{fig: GPT_rate}
\end{figure}
\subsection{High Quality Synthetic Dataset}\label{sec: Synthetic_Dataset}
\begin{table*}[ht]
\centering
\caption{The results on the MME benchmark demonstrate that using generated data can still maintain a leading performance in real-world problems, further expanding the application scope of SynthVLM.}
\resizebox{1.0\textwidth}{!}{
\begin{tabular}{@{}cc|ccc|cccccc@{}} 
\toprule 

\multirow{2}{*}{\textbf{Models}} & \multirow{2}{*}{\textbf{LLM}} & \multicolumn{3}{c|}{\textbf{MME Cognition}} & \multicolumn{6}{c}{\textbf{MME Perception}} \\ 
\cmidrule(lr){3-11} 
& & \textbf{Reasoning} & \textbf{Translation} & \textbf{Code} & \textbf{Posters} & \textbf{Celebrity} & \textbf{Scene} & \textbf{Landmark} & \textbf{Artwork} & \textbf{OCR} \\
\midrule 
\textbf{LLaVA1.5-7B} & Vicuna-1.5-7B & 126.4 & 57.5 & \textbf{62.5} & 148.3 & \textbf{132.1} & 143.0 & 141.8 & 123.8 & 100.0 \\
\textbf{SynthVLM-7B} & Vicuna-1.5-7B & \textbf{143.6} & \textbf{70.0} & 57.5 & \textbf{157.1} & 124.2 & \textbf{157.8} & \textbf{156.5} & \textbf{136.2} & \textbf{102.5} \\
\midrule
\textbf{LLaVA-13B} & Vicuna-1.5-13B & 119.3 & 50.0 & 62.5 & 155.4 & 127.4 & 158.5 & \textbf{165.2} & 129.3 & \textbf{110.0} \\
\textbf{SynthVLM-13B} & Vicuna-1.5-13B & \textbf{130.0} & \textbf{65.0} & 62.5 &  \textbf{160.4} & \textbf{138.2} & \textbf{162.0} & 158.8 & \textbf{140.3} & 104.5 \\
\bottomrule 
\end{tabular}
}
\label{tab: Main_3}
\end{table*}

In this section, we compare commonly used image-caption datasets with the SynthVLM-100K dataset. The synthetic data offers high image quality, excellent image-text alignment and superior model ratings.
\paragraph{\textbf{High Image Quality}}
As illustrated in Figure \ref{fig: Face_1}, SynthVLM markedly improves image quality by generating images at a resolution of 1024x1024 pixels. This high resolution addresses the prevalent issue of suboptimal image quality in existing datasets, providing high-quality image-caption pairs that are invaluable for training VLMs. Additionally, SynthVLM-100K effectively eliminates artifacts such as watermarks and advertisements. 

\paragraph{\textbf{Excellent Image-Text Alignment.}}  
As shown in Table \ref{table: CLIPScore}, the SynthVLM-100K dataset exhibits a higher CLIPScore and SSIMScore compared to existing high-quality web datasets. By selecting curated image-text pairs of superior quality, SynthVLM-100k surpasses datasets such as COCO-Caption, BLIP-LCS, and ShareGPT4V. This demonstrates the outstanding alignment of our dataset.

\paragraph{\textbf{Excellent Model Rating.}} Since our data will be used for VLMs training, we use VLMs to evaluate the data quality. We selected 1K image-caption pairs and submitted the caption along with the synthetic image and the original image. We used human annotators, GPT-4 Vision\cite{gpt4} and Intern-VL2\cite{internvl2} as the judge model and requested it to select the pair that exhibited higher alignment. The specific prompt used for this evaluation is illustrated in Figure \ref{fig: GPT_rate}. The results, presented in Table \ref{table: GPT_rate}, demonstrate that images generated have better alignment with the caption.

\begin{table}
\caption{We employed GPT4-Vision and InternVL to vote on the match between each caption and its corresponding generated image and raw image. The results demonstrate that the generated images align more closely with the captions.}
\resizebox{0.47\textwidth}{!}{
    \centering
    \begin{tabular}{cc|cc} 
    \toprule
    \textbf{Sample} & \textbf{Model} & \textbf{Sythetic Image win} & \textbf{Web Image win} \\
    \midrule
    1K & GPT4-Vision & 633 & 367 \\
    1K & InternVL2 & 692 & 308 \\
    1K & Human & 758 & 242\\
    \bottomrule
    \end{tabular}
}
\label{table: GPT_rate}
\end{table}

Through these two stages, we successfully developed SynthVLM-100K data. Our method is efficient, utilizing only 100K pre-training data. Additionally, SynthVLM provides a new paradigm for effective alignment between modalities in Vision Language Models using pure synthetic data.

\begin{table}[ht]
\centering
\caption{Comparison of data utilization for generating image-caption pairs. This indicates that our SynthVLM have superior efficiency compared to other methods.}
\begin{tabular}{@{}lccc@{}} 
\toprule 
\textbf{Methods} & \textbf{SynthVLM} & \textbf{LLaVA} & \textbf{w/o selection} \\
\midrule 
\textbf{Dataset Number (k)} & 100 & 558 & 1000\\
\textbf{Data Usage} & 33MB & 27GB & 310MB\\
\bottomrule 
\end{tabular}
\label{tab: Efficient}
\end{table}

\section{Experiments}

In this section, we utilize the image data synthesis system for various tasks. We then aim to answer the following questions to verify the effectiveness and efficiency of our proposed SynthVLM: \textbf{Q1}: Can our SynthVLM-7B and SynthVLM-13B achieve SOTA performance compared to previous SOTA methods? \textbf{Q2}: How does the efficiency of our SynthVLM compare to previous methods? \textbf{Q3}: Would using generated data in full impact the application of the model in real-world scenarios? \textbf{Q4}: Do we need the generate module and the quality selection module to enhance model performance?

\subsection{Experimental Settings}
\paragraph{\textbf{Models.}} We use the LLaVA 1.5\cite{llava1.5} model to validate the effectiveness of our dataset. For the Vision Encoder, we select CLIP 336px, and for the LLM, we use Vicuna 7B and Vicuna 13B\cite{vicuna2023}. Our training parameters are consistent with LLaVA, and we adopt a two-stage training approach. In the pre-training stage, we train the projector to align the image and text modalities. The SynthVLM-100K dataset described in Section \ref{sec: Synthetic_Dataset} is used for this purpose. In the SFT stage, we further train the projector alongside the LLM to improve visual understanding capabilities.

\paragraph{\textbf{Datasets.}} For LLaVA-BaseLine, we use the LLaVA 558k dataset for pre-training and the LLaVA 665k dataset for SFT. For SynthVLM-7B and SynthVLM-13B, we use the SynthVLM-100K dataset for pre-training and the LLaVA 665k dataset for SFT.

\paragraph{\textbf{Benchmarks.}} We select benchmarks for both visual understanding and language understanding. For visual understanding, we choose SQA$^I$\cite{sqa}, MMVet\cite{mmvet}, VizWiz\cite{vizwiz}, VQAv2\cite{vqav2}, GQA\cite{gqa}, MME\cite{mme}, and PoPE\cite{pope} for a comprehensive evaluation. For language benchmarks, we select MMLU\cite{mmlu} and SQA\cite{sqa} to assess language understanding abilities.

\begin{table*}[!ht]
\centering
\caption{Ablation study of visual understanding ability and pure language ability. The results demonstrate that removing either the data generation or data selection module results in a performance drop.}
\resizebox{1.0\textwidth}{!}{
    \begin{tabular}{ccccccccccccc}
    \toprule
    \textbf{Models} & \textbf{LLM} & \textbf{SQA} & \textbf{SQA$^I$} & \textbf{MMVet} & \textbf{VizWiz} & \textbf{VQAv2} & \textbf{GQA} & \textbf{MMB} & \textbf{MME$^P$} & \textbf{MME$^C$} & \textbf{PoPE} & \textbf{MMLU} \\
    \midrule
    \textbf{SynthVLM-7B}  & Vicuna-1.5-7B     & 70.4 & 68.9 & 32.2  & 49.3 & 79.4 & 63.1 & 66.8 & 1518.5 & 345.7 & 87.0 & 41.2 \\
    \textbf{w/o generation} & Vicuna-1.5-7B& 69.3\textcolor{blue}{$\downarrow$} & 67.0\textcolor{blue}{$\downarrow$} & 31.2\textcolor{blue}{$\downarrow$} & 46.8\textcolor{blue}{$\downarrow$} & 79.3\textcolor{blue}{$\downarrow$} & 62.9\textcolor{blue}{$\downarrow$} & 66.2\textcolor{blue}{$\downarrow$} & 1488.8\textcolor{blue}{$\downarrow$} & 327.5\textcolor{blue}{$\downarrow$} & 86.2\textcolor{blue}{$\downarrow$} & 39.1\textcolor{blue}{$\downarrow$} \\
    \textbf{w/o selection} & Vicuna-1.5-7B& 69.9\textcolor{blue}{$\downarrow$} & 67.7\textcolor{blue}{$\downarrow$} & 30.2\textcolor{blue}{$\downarrow$}  & 50.2 & 79.1\textcolor{blue}{$\downarrow$} & 62.2\textcolor{blue}{$\downarrow$} & 63.5\textcolor{blue}{$\downarrow$} & 1421.7\textcolor{blue}{$\downarrow$} & 301.8\textcolor{blue}{$\downarrow$} & 87.3 & 40.6\textcolor{blue}{$\downarrow$} \\
    \midrule
    \textbf{SynthVLM-13B}  & Vicuna-1.5-13B     & 74.9 & 72.5 & 35.0  & 55.9 & 80.0 & 63.5 & 68.3 & 1573.0 & 316.1 & 88.4 & 54.6 \\
    \textbf{w/o generation} & Vicuna-1.5-13B& 73.6\textcolor{blue}{$\downarrow$} & 71.4\textcolor{blue}{$\downarrow$} & 33.0\textcolor{blue}{$\downarrow$} & 53.6\textcolor{blue}{$\downarrow$} & 80.0 & 63.4\textcolor{blue}{$\downarrow$} & 67.5\textcolor{blue}{$\downarrow$} & 1514.3\textcolor{blue}{$\downarrow$} & 295.7\textcolor{blue}{$\downarrow$} & 88.2\textcolor{blue}{$\downarrow$} & 53.6\textcolor{blue}{$\downarrow$} \\
    \textbf{w/o selection} & Vicuna-1.5-13B& 74.1\textcolor{blue}{$\downarrow$} & 70.5\textcolor{blue}{$\downarrow$} & 35.6 & 53.2\textcolor{blue}{$\downarrow$} & 79.7\textcolor{blue}{$\downarrow$} & 63.1\textcolor{blue}{$\downarrow$} & 67.5\textcolor{blue}{$\downarrow$} & 1512.7\textcolor{blue}{$\downarrow$} & 303.2\textcolor{blue}{$\downarrow$} & 86.9\textcolor{blue}{$\downarrow$} & 53.0\textcolor{blue}{$\downarrow$} \\
    \bottomrule
    \end{tabular}
}
\label{table: Ablation_1}
\end{table*}

\begin{table}[!ht]
\centering
\caption{Ablation study of modality alignment. The results demonstrate that removing either the data generation or data selection module results in a performance drop.}
\resizebox{0.47\textwidth}{!}{
\begin{tabular}{@{}cccccc|c@{}} 
\toprule 
\multirow{2}{*}{\textbf{Models}} & \multirow{2}{*}{\textbf{SQA}} & \multicolumn{5}{c}{\textbf{MMLU}} \\ 
\cmidrule(lr){3-7} 
& & \textbf{Avg} & \textbf{STEM} & \textbf{Humanities} & \textbf{Social Sciences} & \textbf{Other} \\
\midrule 
\textbf{SynthVLM-7B}  & 70.4 & 41.2 & 31.7 & 37.4 & 47.0 & 50.2 \\
\textbf{w/o generation}  & 69.3\textcolor{blue}{$\downarrow$} & 39.1\textcolor{blue}{$\downarrow$} & 30.0\textcolor{blue}{$\downarrow$} & 36.6\textcolor{blue}{$\downarrow$} & 43.1\textcolor{blue}{$\downarrow$} & 47.3\textcolor{blue}{$\downarrow$} \\
\textbf{w/o selection}  & 69.9\textcolor{blue}{$\downarrow$} & 40.6\textcolor{blue}{$\downarrow$} & 30.8\textcolor{blue}{$\downarrow$} & 37.2\textcolor{blue}{$\downarrow$} & 45.3\textcolor{blue}{$\downarrow$} & 48.9\textcolor{blue}{$\downarrow$} \\
\midrule
\textbf{SynthVLM-13B}  & 74.9 & 54.6 & 45.0 & 49.3 & 64.0 & 62.2 \\
\textbf{w/o generation} & 74.1\textcolor{blue}{$\downarrow$} & 53.6\textcolor{blue}{$\downarrow$} & 43.5\textcolor{blue}{$\downarrow$} & 48.2\textcolor{blue}{$\downarrow$} & 63.1\textcolor{blue}{$\downarrow$} & 61.8\textcolor{blue}{$\downarrow$} \\
\textbf{w/o selection}  & 73.6\textcolor{blue}{$\downarrow$} & 53.0\textcolor{blue}{$\downarrow$} & 42.9\textcolor{blue}{$\downarrow$} & 46.8\textcolor{blue}{$\downarrow$} & 63.8\textcolor{blue}{$\downarrow$} & 61.3\textcolor{blue}{$\downarrow$} \\
\bottomrule 
\end{tabular}
}
\label{table: Ablation_2}
\end{table}

\paragraph{\textbf{Settings.}}
We primarily adhered to the hyperparameters specified in the official repository for the LLaVA model evaluations. All experiments were conducted on an 8x NVIDIA A100 GPU machine equipped with a 120-core CPU and 960GB of memory.

\subsection{Synthetic Data Achieves SOTA Performance}
To address \textbf{Q1}, we trained the model described in experimental settings. From Table \ref{tab: Main_1}, it is evident that SynthVLM model outperforms the Baseline across all evaluation benchmarks on both 7B and 13B. SynthVLM model also excels in language benchmarks, demonstrating superior performance in SQA and MMLU, thus showcasing its comprehensive capabilities in both vision and language tasks.

These results demonstrate the strong alignment capability of our synthetic data. Additionally, this provides a new paradigm for effective visual understanding model modality alignment using generated data. During pre-training, it is common to train on all available data due to uncertainty about data selection. Here, we offer 100k high-quality synthetic data as a benchmark for selecting aligned generated data efficiently.

\subsection{Efficient Vision Language Alignment}
To address \textbf{Q2}, we examine the computational resource usage during training and evaluate the data utilization efficiency for generating image-caption pairs.

As shown in Figure ~\ref{tab: Efficient}, by integrating a data selection module, our approach utilizes only 19\% of the LLAVA data and 10\% of the original synthetic data while achieving SOTA performance. This demonstrates that our data selection method can reduce computational usage by more than 80\%.

\subsection{Capability to Solve Real-World Problems}
A critical issue when using generated data is whether the model loses its ability to solve real-world problems. To address \textbf{Q3}, we utilized the MME Benchmark to assess the model’s performance in various real-world scenarios, such as artwork, celebrity, code reasoning, landmarks, and posters. The specific results are presented in Table~\ref{tab: Main_3}. Using SynthVLM-100K, our model performs consistently well, matching or even surpassing the baseline across most real-world tasks. We attribute this success to the diversity of the captions and the generalization ability of the Diffusion model. 

\subsection{Ablation Study}
To address \textbf{Q4}, We conducted an ablation study where we removed the data generation module and the data selection module separately to evaluate their individual contributions to the effectiveness of our data generation pipeline. In this section, we controlled the number of samples in each experimental group to be 100K.

\paragraph{\textbf{Excluding Data Generation Module}}
The exclusion of the data generation module significantly impacts the model's performance, as illustrated in Tables \ref{table: Ablation_1} and \ref{table: Ablation_2}, labeled as "w/o generation". The variant without this module demonstrates lower accuracy, emphasizing the crucial role of the data generation process in sustaining the high performance of the SynthVLM model. This also underscores SynthVLM's potential in constructing highly aligned datasets.

\paragraph{\textbf{Excluding Data Selection Module}} 
The absence of the data selection module similarly leads to a noticeable decline in performance, indicated as "w/o selection" in Tables \ref{table: Ablation_1} and \ref{table: Ablation_2}. Given the inherent randomness of diffusion models, which inevitably generate some low-quality images, the data selection module is crucial for removing these subpar elements.

Overall, the ablation study highlights the critical role of data generation and data selection in SynthVLM, providing valuable insights into the contributions of each module.

\section{Conclusion}
We propose SynthVLM, a novel pipeline for generating high-quality pre-training data for VLMs. Unlike existing datasets, our synthetic images are free from watermarks and advertisements, leading to cleaner visual inputs. The generated data demonstrate superior alignment and visual fidelity. Notably, the SynthVLM model—trained with only 18\% synthetic data—outperforms the LLaVA baseline trained on the full dataset. It not only achieves state-of-the-art alignment performance but also preserves the language understanding capabilities of VLMs. Ablation studies further confirm the effectiveness of both the image generation and data selection modules, highlighting the practical value of our data synthesis approach.

\section*{Acknowledgments}
This work is supported by the National Key R\&D Program of China (2024YFA1014003), National Natural Science Foundation of China (92470121, 62402016), and High-performance Computing Platform of Peking University.

\bibliographystyle{ACM-Reference-Format}
\bibliography{main}

\appendix
\clearpage
\setcounter{page}{1}
\makeatletter
\newcommand{\thetitle}{\@title}
\makeatother

\newcommand{\maketitlesupplementary}{%
  \begin{center}
    {\Huge\bfseries \thetitle\par}
    \vspace{0.5em}
    \Large Appendix\par
    \vspace{1.0em}
  \end{center}%
}
\twocolumn[{%
\renewcommand\twocolumn[1][]{#1}%
\maketitlesupplementary
}]

\section{Preliminary}
\subsection{Diffusion Model}
Denoising diffusion probabilistic models (DDPMs) ~\cite{ddpm, ldm, sdxl} are a class of generative models renowned for their ability to generate extremely high-quality images. The core idea of DDPMs involves modeling the data distribution by gradually adding Gaussian noise to the input image during the forward process and then predicting and removing this noise to reconstruct the image during the backward process.

Given a source image data distribution \( x_0 \sim q(x_0) \), Gaussian noise is added over \( T \) steps to obtain \( x_T \). The forward process is defined as:
\[
q(x_1, \ldots, x_T \mid x_0) := \prod_{t=1}^{T} q(x_t \mid x_{t-1}),
\]
\[
q(x_t \mid x_{t-1}) = \mathcal{N}(x_t; \sqrt{1 - \beta_t} x_{t-1}, \beta_t I),
\]
where \( \beta_t \) controls the variance of the noise added at each step.

The distribution after \( t \) steps can be written as:
\[
q(x_t \mid x_0) = \mathcal{N}(x_t; \sqrt{\bar{\alpha}_t} x_0, (1 - \bar{\alpha}_t) I),
\]
where \( \bar{\alpha}_t = \prod_{i=1}^t (1 - \beta_i) \).

The backward process aims to reconstruct the data by learning a series of Gaussian distributions that approximate the forward process:
\[
p_\theta(x_{t-1} \mid x_t) = \mathcal{N}(x_{t-1}; \mu_\theta(x_t, t), \Sigma_\theta(x_t, t)),
\]
where \( \mu_\theta \) and \( \Sigma_\theta \) are neural networks parameterized by \( \theta \).

While DDPMs have shown promising results, several improvements have been proposed to enhance their efficiency ~\cite{ddim, asdm} and sample quality ~\cite{iddpm, dmgans}. The superior performance of diffusion models has been leveraged in various sub-tasks, including image generation, image translation, inpainting ~\cite{dm2speech, dualdm}. 
\subsection{Vision Language Models}
The integration of visual knowledge into large language models (LLMs) has become a pivotal area of research due to the rapid advancements in LLMs. VLMs combine vision information from vision encoders with LLMs, thus enabling these models to process and interpret visual inputs for various visual tasks~\cite{dino,glipv2, grounded-pt} with enhanced accuracy and efficiency. Pioneering frameworks like CLIP~\cite{clip} leverage contrastive learning on expansive image-caption datasets to align modalities, forming the groundwork for cross-modal comprehension. Various adapters~\cite{llava, llava1.5,  blip, blip1, pformer, lyrics} are introduced to further integrate different modalities. For example, LLaVA~\cite{llava, llava1.5} employs a straightforward MLP to inject the vision information into LLMs. Whereas more complex implementations like the Q-Former in BLIP~\cite{blip1, blip} utilize cross-attention to enhance modality integration. 

Recent studies ~\cite{image-text-data, sharegpt4v, llava, llava1.5, otter} aims to boost VLM performance by focusing on the quality of both pre-training and fine-tuning datasets. Models like LLaVA ~\cite{llava, llava1.5} and ShareGPT4V ~\cite{sharegpt4v} have shown remarkable advancements in understanding and following complex instructions through instruction tuning. Although these improvements help align the vision modality and establish a solid basis for cross-modal comprehension, they require extensive datasets for training and could potentially diminish the model's language capabilities. 

\section{Implementation Details}
\subsection{Data Generation}
In this section, we detail the hyperparameters and procedures used for data generation. 

We employed the Stable Diffusion XL(SDXL) model for image synthesis, following the framework outlined by the original authors~\cite{sdxl}. To identify the optimal parameter configuration for our use case, we conducted a grid search strategy aimed at maximizing the CLIPScore for evaluating the semantic alignment between generated images and their corresponding textual descriptions.

Specifically, we randomly sampled 1k captions from our caption pool and used these samples to systematically evaluate different combinations of generation parameters. The grid search allowed us to empirically determine the most effective configuration for producing high-quality, semantically relevant synthetic images.

Based on this optimization process, we configured SDXL with 60 sampling steps. All images were generated at a resolution of 1024×1024 pixels. These configurations consistently yielded superior quality. 
\subsection{Data Selection}
\begin{table*}[htbp]
    \centering
    \caption{Metric and Prompt used for Caption Filtering}
    \scalebox{1.0}{
    \begin{tcolorbox}[colback=gray!5,colframe=black!75, title=Caption Filtering]
    \small
    \#\# Rule-Based Metrics \\
    \begin{itemize}
    \item \textbf{Alphanumeric Filter:} Tokenization: false, Min ratio: 0.60
    \item \textbf{Character Repetition Filter:} Rep length: 10, Max ratio: 0.09373663
    \item \textbf{Flagged Words Filter:} Language: en, Tokenization: false, Max ratio: 0.0
    \item \textbf{Perplexity Filter:} Language: en, Max perplexity: 5500.0
    \item \textbf{Special Characters Filter:} Min ratio: 0.16534802, Max ratio: 0.42023757
    \item \textbf{Word Repetition Filter:} Language: en, Tokenization: false, Rep length: 10, Max ratio: 0.03085751
    \item \textbf{Image-Text Matching Filter:} HF BLIP: Salesforce/blip-itm-base-coco, Min score: 0.8, Max score: 1.0, Horizontal flip: false, Vertical flip: false, Reduce mode: avg, Any or all: any, Mem required: 1500MB
    \item \textbf{Image-Text Similarity Filter:} HF CLIP: openai/clip-vit-base-patch32, Min score: 0.28
    \end{itemize}
    \vspace{1em}
    \#\# Prompt \\
    Assume you are an expert in the field of AI image generation. Your goal is to select high-descriptive prompts that will enable the successful generation of images. I will provide you with a specific descriptive prompt, and your task is to evaluate it thoroughly. Consider the prompt's level of detail, its logical coherence, and the clarity with which it describes the desired image. It is essential to assess whether the prompt contains sufficient information to guide the diffusion model effectively, ensuring that it can produce an image that meets expectations. You should only respond with Yes or No.
    \end{tcolorbox}
    }
    \label{tab:Caption Filtering}
\end{table*}

In this section, we describe the strategy and prompts used for data selection. Our goal was to curate a high-quality dataset that aligns closely with our generation objectives.

To achieve this, we employed a two-stage filtering process combining heuristic rules and large language model (LLM)-based evaluation. The specific filtering rules and prompt templates are detailed in Table \ref{tab:Caption Filtering}.

For heuristic filtering, we utilized the Data-Juicer framework, which offers a modular and scalable pipeline for rule-based data preprocessing. This allowed us to implement filters targeting criteria such as minimum caption length, syntactic completeness, and lexical diversity. Additionally, we removed low-information and repetitive entries to enhance the overall quality of the dataset.

Following this, we conducted LLM-based filtering using LLaMA3-70B-Instruct, a powerful instruction-tuned language model. This model was used to assess the semantic clarity, descriptiveness, and relevance of each caption to ensure alignment with our image generation goals. Captions that met the predefined criteria for specificity, visual richness, and informativeness were retained.

\section{Another Advantage: Addressing Data Privacy}
Utilizing web-sourced data introduces numerous security and privacy concerns~\cite{privacy, privacysurvey}. They may contain personal information or copyrighted materials, posing potential legal and ethical challenges. Moreover, the inclusion of sensitive or inappropriate content within training datasets can instigate ethical issues, thereby compromising the models' integrity and fairness.

Our synthetic approach removes reliance on real-world personal data (e.g., user photos), safeguarding user privacy throughout the data generation process while maintaining model capability.

\begin{figure}[t]
\centering 
\includegraphics[width=0.47\textwidth]{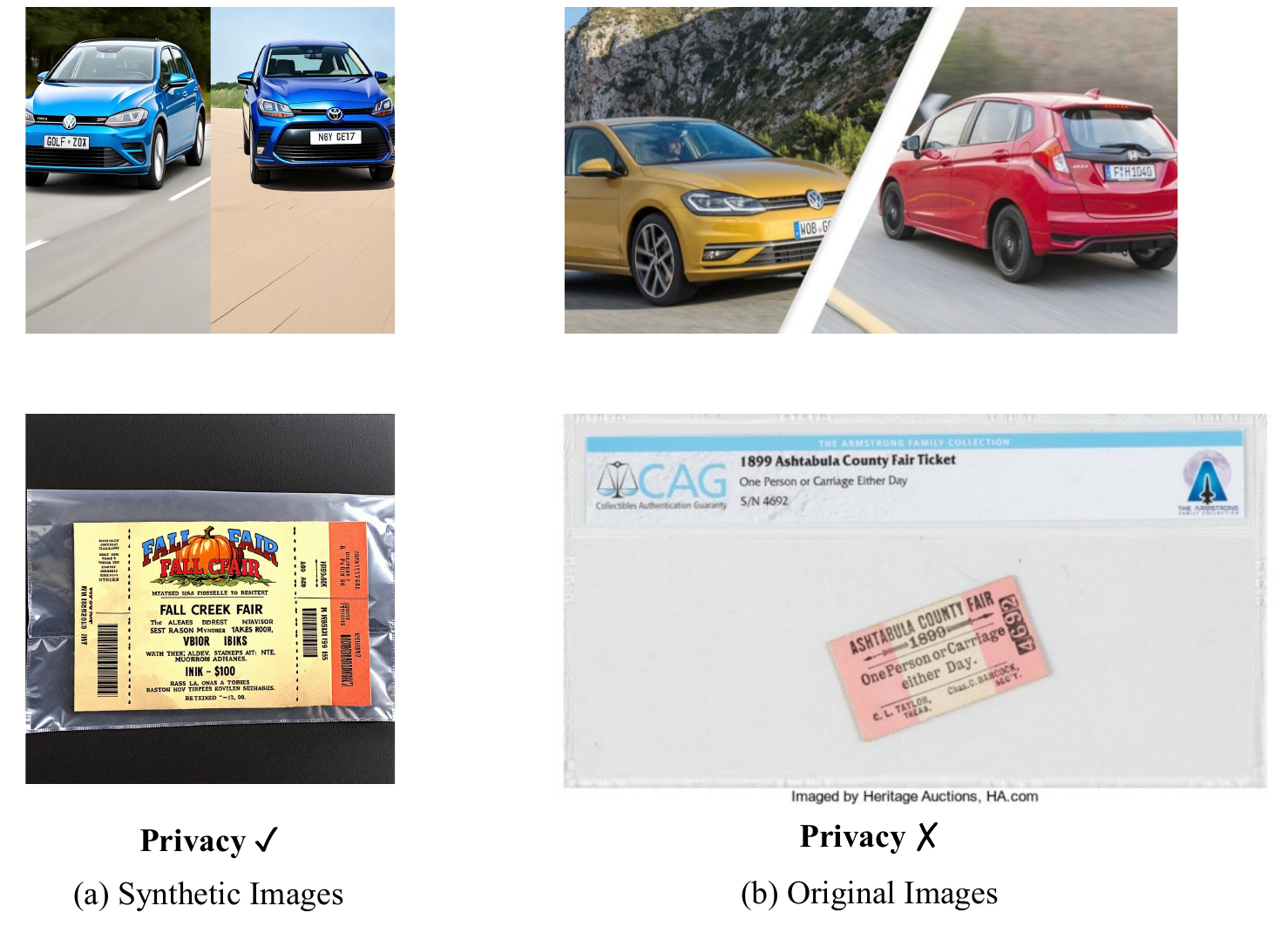} 
\caption{From (a), it is evident that synthetic images can avoid displaying real license plates and ticket information. In contrast, (b) contains actual license plates and ticket information, which can potentially lead to privacy issues.}
\label{fig: Privacy_2}
\end{figure}

we compare the synthetic image and the original image in Figure \ref{fig: Privacy_2}. Synthetic data offers significant advantages in protecting data privacy. In Figure \ref{fig: Privacy_2}, synthetic images in (a) show vehicles and tickets without revealing real license plates and ticket information, ensuring privacy protection. In contrast, original images in (b) display actual license plates and ticket information, which can potentially lead to privacy issues.

\section{T-SNE visualize of our dataset}
\begin{figure*}
\centering
\begin{subfigure}{0.48\textwidth}
    \centering
    \includegraphics[width=\textwidth]{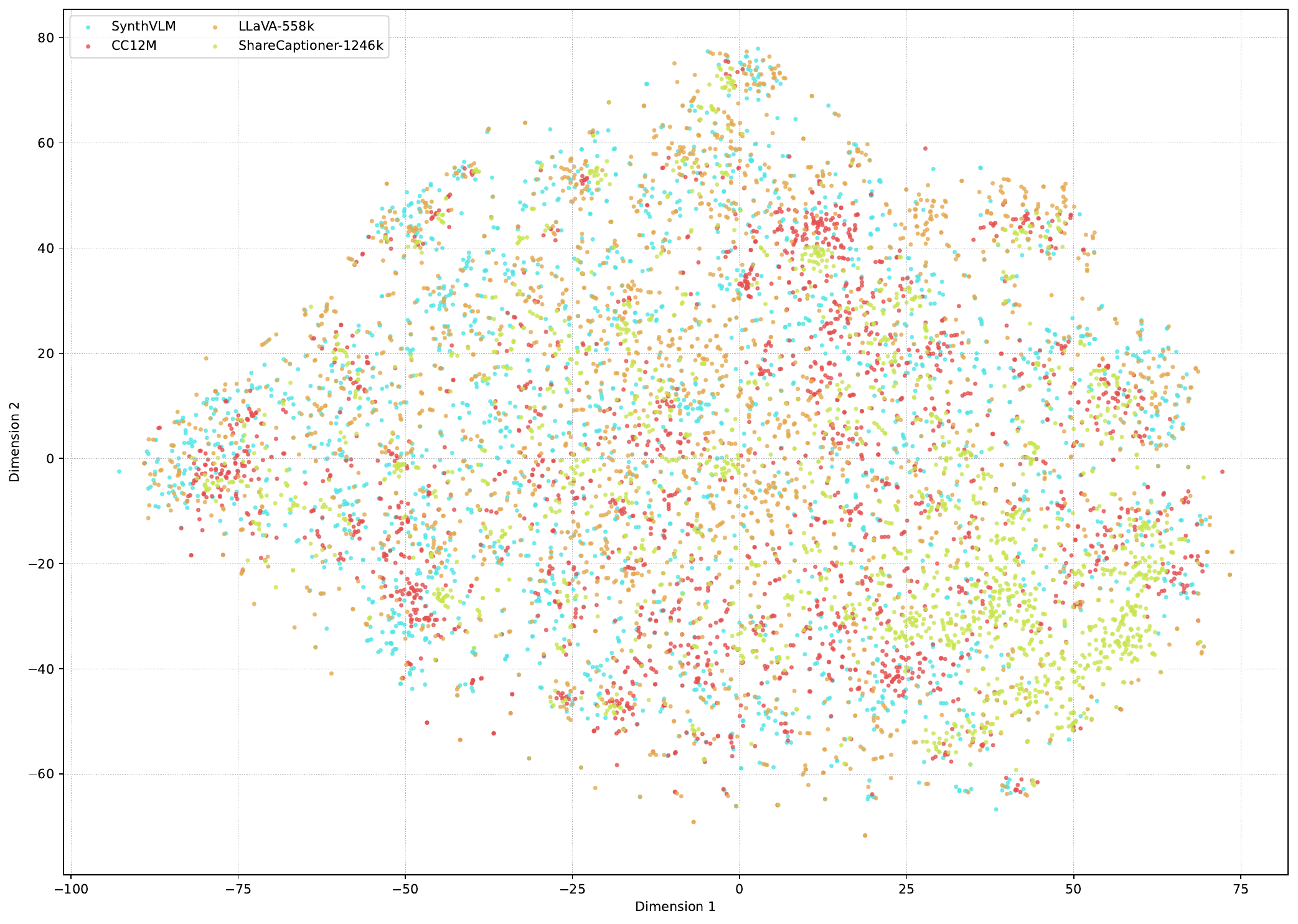} 
    \caption{Visualization of text embeddings.}
    \label{fig:tsne_text_finetune}
\end{subfigure}
\hfill
\begin{subfigure}{0.48\textwidth}
    \centering
    \includegraphics[width=\textwidth]{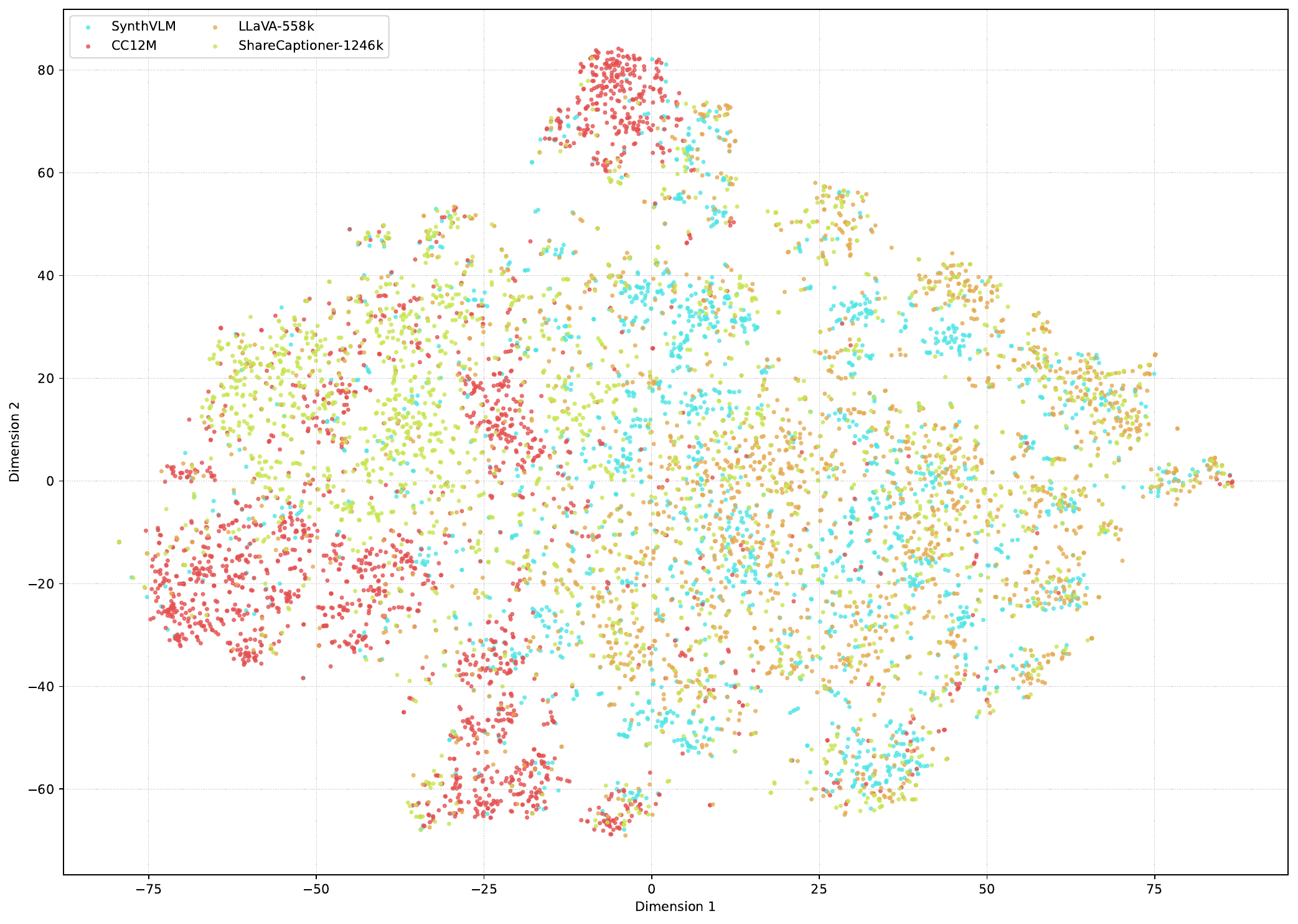}
    \caption{Visualization of image embeddings.}
    \label{fig:tsne_image_finetune}
\end{subfigure}
\caption{TSNE visualizations of synthetic and real datasets for text and image modalities.}
\label{fig:tsne_finetune}
\end{figure*}
In this section, we use t-distributed Stochastic Neighbor Embedding (t-SNE) to compare the distribution of our synthetic dataset with several real-world datasets. This comparison aims to evaluate the similarity in semantic and visual space, providing insights into the realism and utility of the generated data.

For the real datasets, we selected LLaVA-558K\cite{llava1.5}, ShareCaptioner\cite{sharegpt4v}, and CC12M\cite{singla2024pixelsproselargedataset}, which are widely used  for vision-language training. As our synthetic dataset, we used SynthVLM-100K, generated using the methods described in earlier sections. From each dataset, we randomly sampled 1k image-caption pairs for analysis.

We performed t-SNE visualization separately on image features and caption embeddings. Feature representations were extracted using a pre-trained vision-language model to ensure consistency and comparability across datasets.

As shown in Figure \ref{fig:tsne_finetune}, the image and caption distributions of our synthetic dataset are closely aligned with those of the real datasets. This visual overlap indicates that the generated data captures similar semantic and visual characteristics as real-world data, supporting the authenticity and high quality of our generation pipeline.

Furthermore, the observed distributional similarity suggests that models trained on our synthetic data are likely to exhibit strong generalization and performance on real-world tasks. This supports the viability of using synthetic data to supplement or replace real data in various vision-language applications.
\section{More examples of our dataset}
In this section, we present additional qualitative examples from our synthetic dataset, SynthVLM-100K, to further demonstrate the high quality and diversity of the generated image-caption pairs.

As illustrated in Figures 1 through 4, the samples cover a wide range of visual concepts and exhibit strong semantic alignment between images and captions. These examples highlight the capability of our data generation pipeline to produce visually coherent and semantically rich content across various domains.

\begin{figure*}[ht]
\centering
\includegraphics[width=0.85\textwidth]{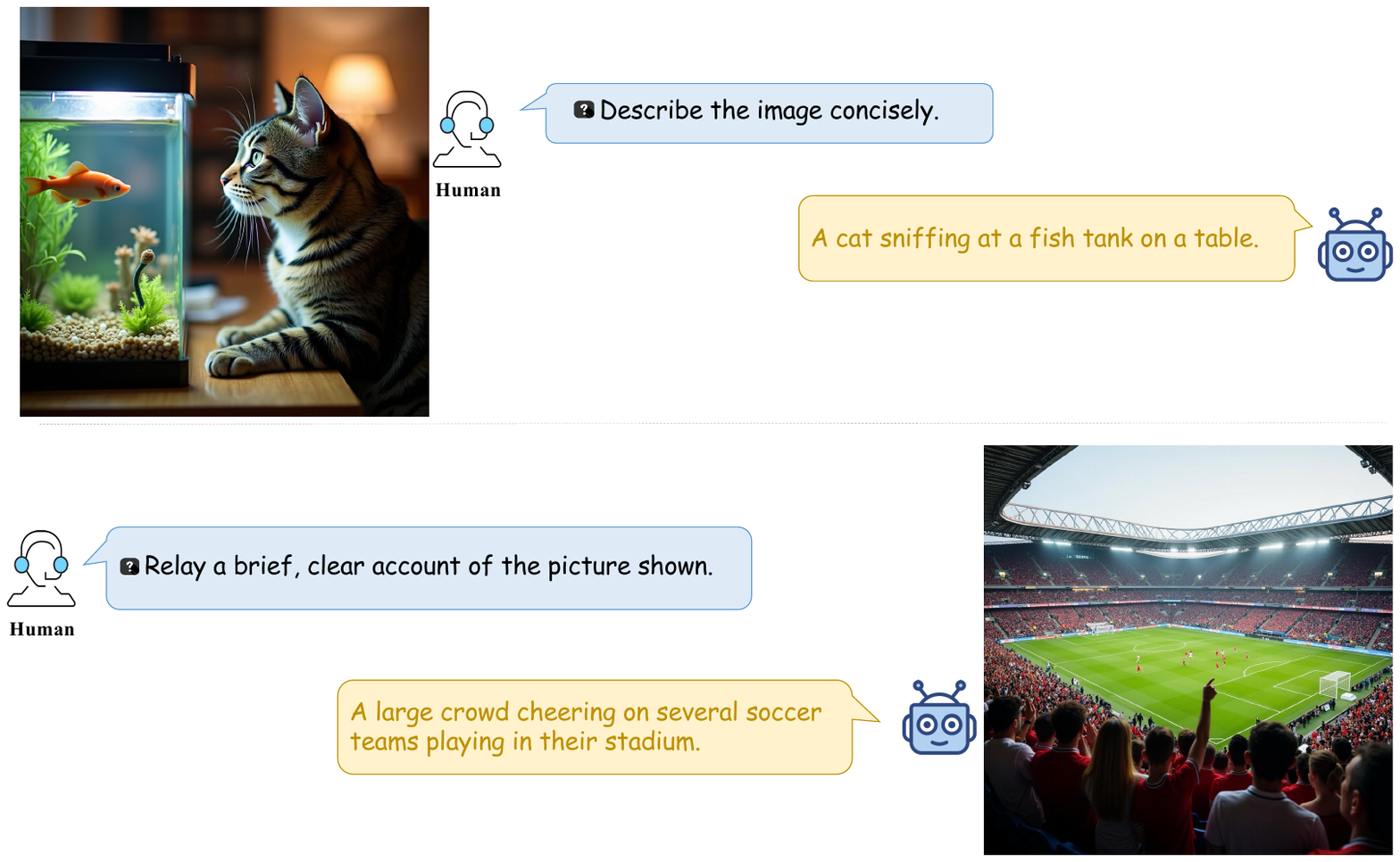}
\vspace{-3em}
\caption{Illustration of our SynthVLM}
\label{fig:example1}
\end{figure*}
\begin{figure*}[ht]
\centering
\includegraphics[width=0.85\textwidth]{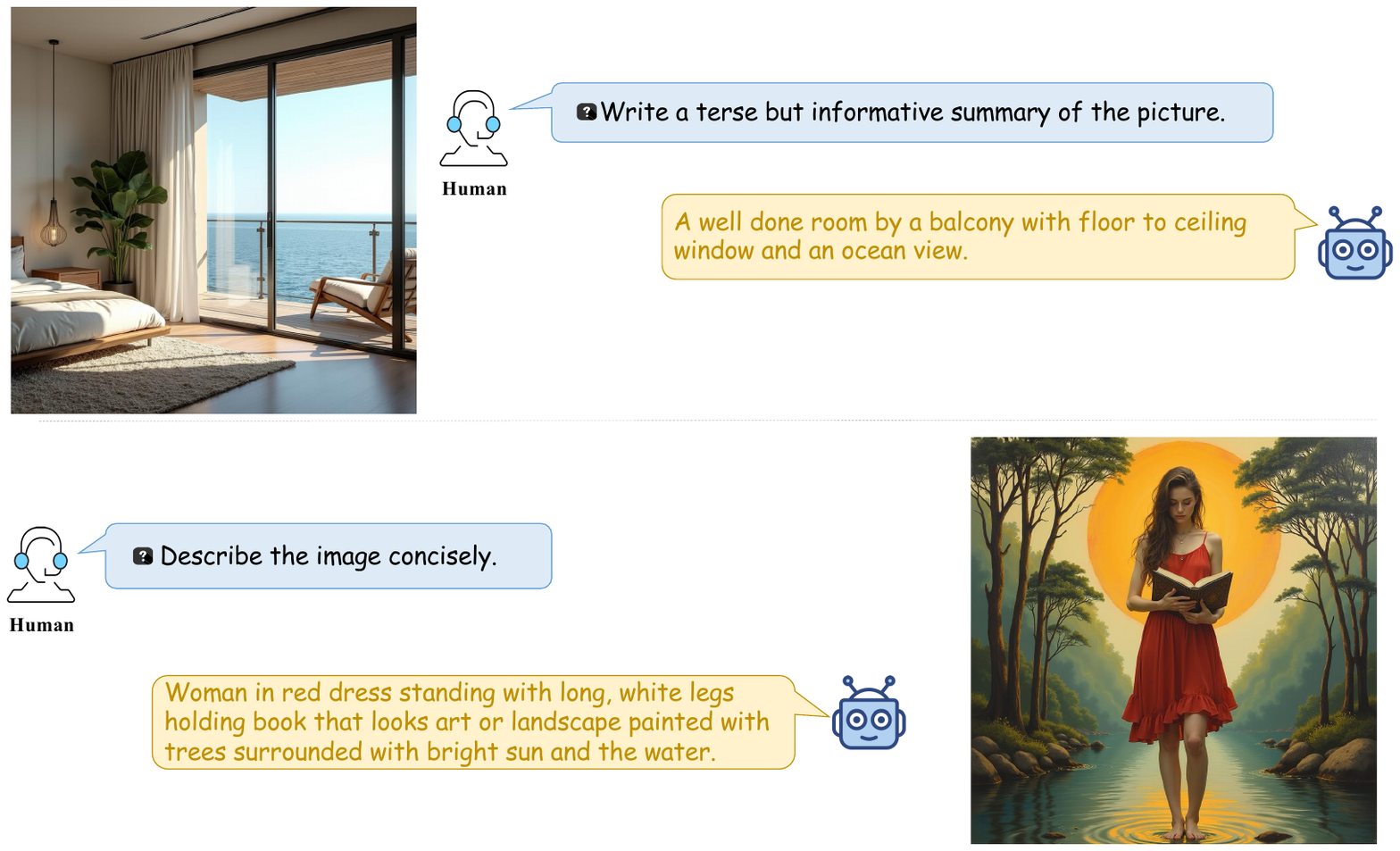}
\vspace{-3em}
\caption{Illustration of our SynthVLM}
\label{fig:example2}
\end{figure*}
\begin{figure*}[ht]
\centering
\includegraphics[width=0.85\textwidth]{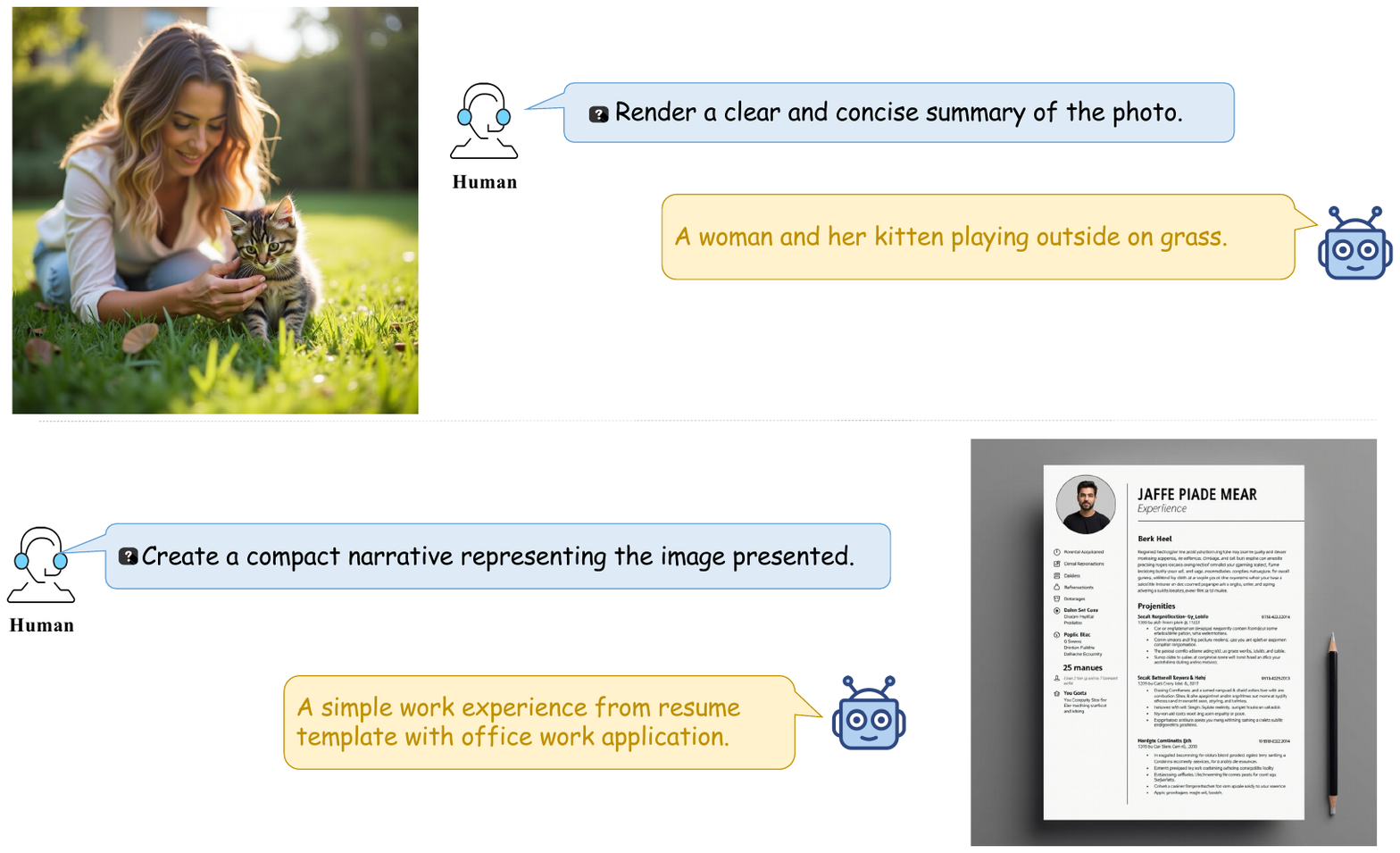}
\vspace{-3em}
\caption{Illustration of our SynthVLM}
\label{fig:example3}
\end{figure*}
\begin{figure*}[ht]
\centering
\includegraphics[width=0.85\textwidth]{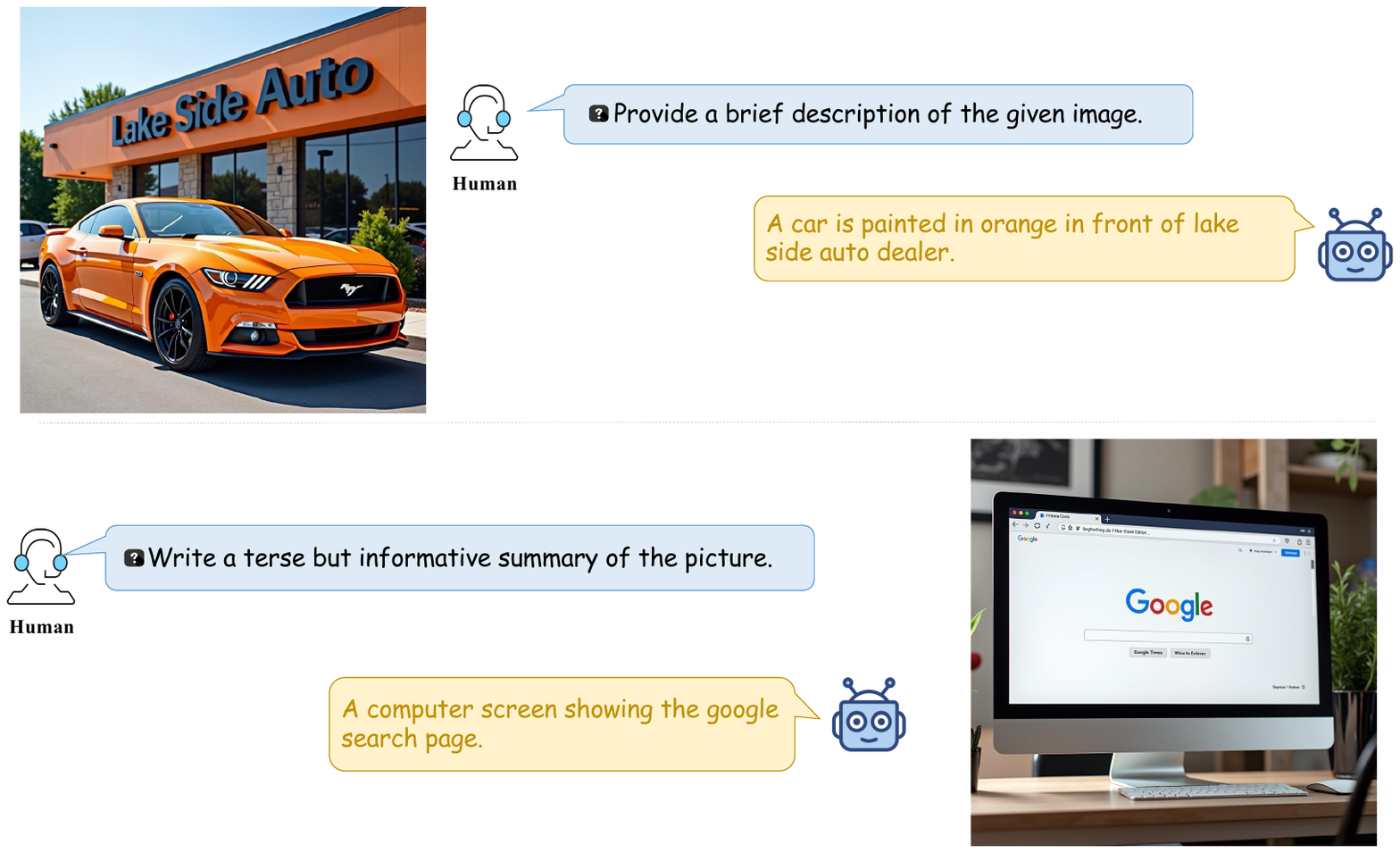}
\vspace{-3em}
\caption{Illustration of our SynthVLM}
\label{fig:example4}
\end{figure*}

\end{document}